\documentclass[a4paper,twoside]{article}

\usepackage{epsfig}
\usepackage{subcaption}
\usepackage{calc}
\usepackage{amssymb}
\usepackage{amstext}
\usepackage{amsmath}
\usepackage{amsthm}
\usepackage{multicol}
\usepackage{pslatex}
\usepackage{apalike}
\usepackage{url}
\usepackage[bottom]{footmisc}
\usepackage{acronym}
\usepackage{todonotes}
\usepackage{array}
\usepackage{graphicx}
\usepackage{float}
\usepackage{caption}
\usepackage{subcaption}

\usepackage{ctable} 
\usepackage{SCITEPRESS}     
\usepackage[outdir=./]{epstopdf}

\newacro{IRS}[IRS]{information retrieval system}
\newacroplural{IRS}[IRS's]{information retrieval systems}
\newacro{ML}[ML]{machine learning}
\newacro{QA}[QA]{question answering}
\newacro{NLP}[NLP]{natural language processing}
\newacro{RLHF}[RLHF]{Reinforcement Learning from Human Feedback}
\newacro{LLM}[LLM]{Large Language Model}
\newacroplural{LLM}[LLM's]{Large Language Models}
\newacro{OCR}[OCR]{optical character recognition}
\newacro{NER}[NER]{Named Entity Recognition}
\newacro{EM}[EM]{Exact Match}
\newacroplural{EM}[EM's]{Exact Matches}
\newacro{ROUGE}[ROUGE]{Recall-Oriented Understudy for Gisting Evaluation}

\newcommand{\eg}{e.g.,~}
\newcommand{\ie}{i.e. }  

\newcommand{\specialcell}[2][c]{%
  \begin{tabular}[#1]{@{}c@{}}#2\end{tabular}}
\newcolumntype{P}[1]{>{\centering\arraybackslash}p{#1}}
\newcolumntype{M}[1]{>{\centering\arraybackslash}m{#1}}
\newcolumntype{B}[1]{>{\centering\arraybackslash}b{#1}}

\DeclareSymbolFont{bbold}{U}{bbold}{m}{n}
\DeclareSymbolFontAlphabet{\mathbbold}{bbold}

\begin{document}

\title{Fine-tuning and aligning question answering models for complex information extraction tasks}

\author{\authorname{
Matthias Engelbach\sup{1}\orcidAuthor{0000-0001-6578-9506}, Dennis Klau\sup{2}\orcidAuthor{0000-0003-3618-7359} and
Felix Scheerer\sup{2}\orcidAuthor{0009-0003-2286-642X} and
Jens Drawehn\sup{1} and
Maximilien Kintz\sup{1}\orcidAuthor{0009-0002-4648-4111}
}
\affiliation{\sup{1}Fraunhofer Institute for Industrial Engineering IAO, Nobelstr. 12, 70569 Stuttgart, Germany}
\affiliation{\sup{2}University of Stuttgart, Institute of Human Factors and Technology Management IAT, Allmandring 35, Stuttgart, Germany}
\email{\{matthias.engelbach, jens.drawehn, maximilien.kitz\}@iao.fraunhofer.de, \{dennis.klau, felix.scheerer\}@iat.uni-stuttgart.de}
}


\abstract{The emergence of Large Language Models (LLMs) has boosted performance and possibilities in various NLP tasks. While the usage of generative AI models like ChatGPT opens up new opportunities for several business use cases, their current tendency to hallucinate fake content strongly limits their applicability to document analysis, such as information retrieval from documents. In contrast, extractive language models like question answering (QA) or passage retrieval models guarantee query results to be found within the boundaries of an according context document, which makes them candidates for more reliable information extraction in productive environments of companies. In this work we propose an approach that uses and integrates extractive QA models for improved feature extraction of German business documents such as insurance reports or medical leaflets into a document analysis solution. We further show that fine-tuning existing German QA models boosts performance for tailored extraction tasks of complex linguistic features like damage cause explanations or descriptions of medication appearance, even with using only a small set of annotated data. 
Finally, we discuss the relevance of scoring metrics for evaluating information extraction tasks and deduce a combined metric from Levenshtein distance, F1-Score, Exact Match and ROUGE-L to mimic the  assessment criteria from human experts.
}

\onecolumn \maketitle \normalsize \setcounter{footnote}{0} \vfill

\section{\uppercase{Introduction}}
\label{sec:introduction}
Automated feature extraction from text documents is a necessary first step for the successful application of many business processes. Unstructured text data needs to be analyzed and stored in structured databases in order to be checked and processed by downstream systems. This task is common to many business areas, \eg customer service centers responding to support requests, insurance companies assessing damage claims or medical authors 
reviewing scientific literature to prepare documents for drug approval procedures -- to name a few examples.

The development of \acp{IRS} supporting in these tasks have a long history~\cite{6182576}. Typically these kind of systems combine capabilities to support different input formats (to deal with scanned as well as electronic text documents) using rules and models for feature extraction. However, recent progress in the field of \acp{LLM} has boosted capabilities of possible applications for \ac{NLP}~\cite{zhang2023preliminary}.

Retrieving some specific information from documents can be arbitrarily complex as text features may appear in form of free wording over several whole sentences (\eg the cause of a cars damage in a damage report which might be given in form of one or several whole sentences).
These kinds of features are difficult to define with rule-based approaches alone in a general way, especially over different context domains and query formulations\footnote{the way an extraction task is defined in the \ac{IRS}}. This qualifies them as prime candidates for \ac{ML}-based extractions, specifically language models capable of capturing and interpreting the textual contexts within a written document.

In contrast to the emerging powerful generative \acp{LLM} like ChatGPT -- which suffer from hallucination and produce output that is usually hard to verify \cite{bang2023multitask} -- the application of extractive \ac{QA} models turned out to be promising for detecting answers in a specific document context \cite{DBLP:journals/corr/abs-2110-03142,DBLP:journals/corr/abs-2110-06393}. Since these models are usually designed to return text boundaries within the given content as output, they are robust to such failure modes. 

For this reason, we investigate the possibilities of applying and fine-tuning extractive \ac{QA} models integrated in an \ac{IRS} and applying it to German text documents with different features (from one-word entities to complex phrases) and business domains, focusing on the following research questions:
\begin{enumerate}
    \item How can Question-Answering (QA) models be used for extraction of complex information from textual documents in specific industrial use cases?
    \item To what extent does the fine-tuning of QA models influence performance across different domains and textual features?
    \item What metrics are appropriate for (automated) performance evaluations that resemble human expert examination?
\end{enumerate}

The remainder of this is paper is structured as follows: in Section~\ref{section:relatedwork}, we present similar research and approaches in the field. Section~\ref{section:ourapproach} shows our approach for using \ac{QA} models in complex information extraction tasks. In Section~\ref{section:evaluation}, we define evaluation metrics, present our evaluation method and the results. Finally, Section~\ref{section:conclusion} summarizes the main findings and lists improvement ideas we are currently working on.

\section{\uppercase{Related work}}
\label{section:relatedwork}

\begin{table*}[h]
\caption{Classification of typical features during extraction tasks with three levels of feature complexity: Simple, Dynamic and Complex. Each type of feature recommends a different type of extraction methods like rule based or trained ML model. Based on previous work \cite{Engelbach_Klau_Drawehn_Kintz_2022}}
\label{tab:extractiontypes}
\renewcommand{\arraystretch}{1.2}
\begin{tabular*}{\textwidth}{@{\extracolsep{\fill}}|m{3.5cm}|m{5.3cm}|m{5.4cm}|}
  \hline
  \multicolumn{1}{|c|}{\textbf{Classification Difficulty}} & \multicolumn{1}{c|}{\textbf{Features examples}} & \multicolumn{1}{c|}{\textbf{Extractor type}} \\
  \hline
  Simple & IBAN, E-mail address, Postal Codes & Rule-based (\eg regular expressions) \\
  \hline
  Dynamic & Named Entities (\eg organization, person name, place) & Trained extraction models (\eg \ac{NER}) \\
  \hline
  Complex & Cause of an event, name of person with a given role & Question-answering model \\
  \hline
\end{tabular*}
\end{table*}

Analysis and information extraction from business documents consists of many tasks, starting with image analysis and text region detection, OCR and text classification to actual entity extraction. For example Tang et al. has tackled these issues and propose an extensive solution on the basis of Microsoft Azure Document AI technology~\cite{https://www.microsoft.com/en-us/research/project/document-ai/,https://doi.org/10.48550/arxiv.2212.02623}. However, for many use cases dealing with the analysis of confidential or personal data, entirely relying on third-party cloud infrastructure is not a viable option. Many companies require solutions that can be fine-tuned to their specific needs and deployed on premise.

In \cite{Kintz_Dukino_Blohm_Hanussek_2020} we proposed a solution for specific data contexts and tasks in the area of information retrieval from written documents.
The approach describes a system for the management of customer claims, that automatically extracts the key entities for handling a claim by utilizing a set of rule-based functions as well as \ac{NER} models.
In \cite{Engelbach_Klau_Drawehn_Kintz_2022} we followed a similar approach, where an automatic address data extraction framework was built. The work compares the various aspects and consequences of implementing either rule-based or deep learning models in \acp{IRS} under diverse input feature and boundary conditions. Afterwards, an \ac{IRS}  framework is proposed that combines both approaches in a single detection and evaluation pipeline. 
Both solutions reported good results in combining human defined rule sets and \ac{ML} extractors, \eg for \ac{NER}, for confident feature retrieval, or for result validation. However, the extraction pipelines described there rely on standard algorithms and \ac{ML} models and do not profit from (con)text understanding capabilities of modern LLMs.

Previous work on document analysis pipelines with scanned images as input in form of IR products like Transkribus \cite{8270253} has been published as well. However, the system focuses on digitalizing content of historical documents and lacks capabilities for flexible feature extraction and layout handling in modern business documents.

In the popular and continuously growing Huggingface community \cite{wolf-etal-2020-transformers} a lot of language models have been open sourced. Although many multilingual models have been published there, the number of good performing models for German language is still limited. 
In the field of \acf{QA}, the SQUAD dataset is a popular dataset for training extractive \ac{QA} models, which aim to find answers to a query within the given boundaries of a context document \cite{SQuAD_Rajpurkar,SQuAD2}. The GermanQuAD dataset \cite{moller-etal-2021-germanquad} and QA models trained by deepset, the associated company, provides a good starting point for our work. Furthermore, deepset also provides a freely available and locally deployable annotation tool~\cite{haystack} suited for \ac{QA} related tagging that we utilized for the creation of training and test ground truths for our documents and model trainings.

The choice of evaluation metrics is an important parameter in determining the applicability of a model to a specific task. Using a single or combination of metrics that correlate well with human judgement for the presented domain can be a powerful approach to reduce the required labeling effort.
\cite{han2021cushlepor} addressed this issue by defining a customized version of the \textit{hLEPOR} metric \cite{Han2013LanguageindependentMF} and tuning its hyperparameters to match either the output of pre-trained language models or human evaluation data. The metric is designed for and optimized on a general collection of data for the task of neural machine translation (NMT) and shown to be a good alternative to the commonly used \textit{BLEU} metric \cite{bleu2002Papineni} for specific language pairs, while our work focuses on the domain of extractive \ac{QA}. \\
Instead of using metrics, an alternative approach to capture the implicit judgement rules of humans is the now widely applied alignment technique \ac{RLHF} for \acp{LLM} \cite{RLHP2023_Korbak,Ziegler2019FineTuningLM,Sparrow2022_Glaese,InstructGPT2022_Ouyang,lambert2022rlhf}: By training a reward model on a set of \ac{LLM} output and human ranking pairs for a given query and using it to optimize the original \ac{LLM} in a reinforcement learning setting, the language model behavior can be implicitly steered in any desired direction depending on the ranking approach. This, however, requires different label types and significant additional training overhead.

Additionally, \cite{schaeffer2023emergent} pointed out that the choice of metrics is important when evaluating the performance of models with respect to sudden performance jumps (also labeled \textit{emergence}) and the associated perception of the model's capabilities by humans. To circumvent the misconception of too expressive individual metrics, this work uses multiple score in conjunction to evaluate the used \ac{QA} models.

\section{\uppercase{Information Extraction Approach}}
\label{section:ourapproach}

The extraction of features from real documents in different business scenarios can be a challenging task, since the information that needs to be detected within the given contexts can be volatile among different companies, industry domains and document types.


In general, extraction methods face different levels of difficulty, depending on the type of information the system attempts to extract automatically. Based on the former work \cite{Kintz_Dukino_Blohm_Hanussek_2020,Engelbach_Klau_Drawehn_Kintz_2022}
we classify the difficulty levels in three categories with examples about the kind  of information to extract and algorithmic approach usually required to tackle them. The classification is shown in Table~\ref{tab:extractiontypes}.

In our previous work \cite{Engelbach_Klau_Drawehn_Kintz_2022}, we implemented a document processing pipeline that supports all steps to gain structured data results from scanned documents. Furthermore, we introduced a flexible framework for the implementation of different new extractors based on rules, regular expressions or trained machine learning models.
In the following, we describe the application and integration of \ac{QA} models into our analysis pipeline for providing means of complex feature extraction that may be indexed by a text span of phrases or even whole sentences within a document.

\subsection{Information extraction pipeline}
 
 \begin{figure*}[h]
  \centering
   {\epsfig{file = 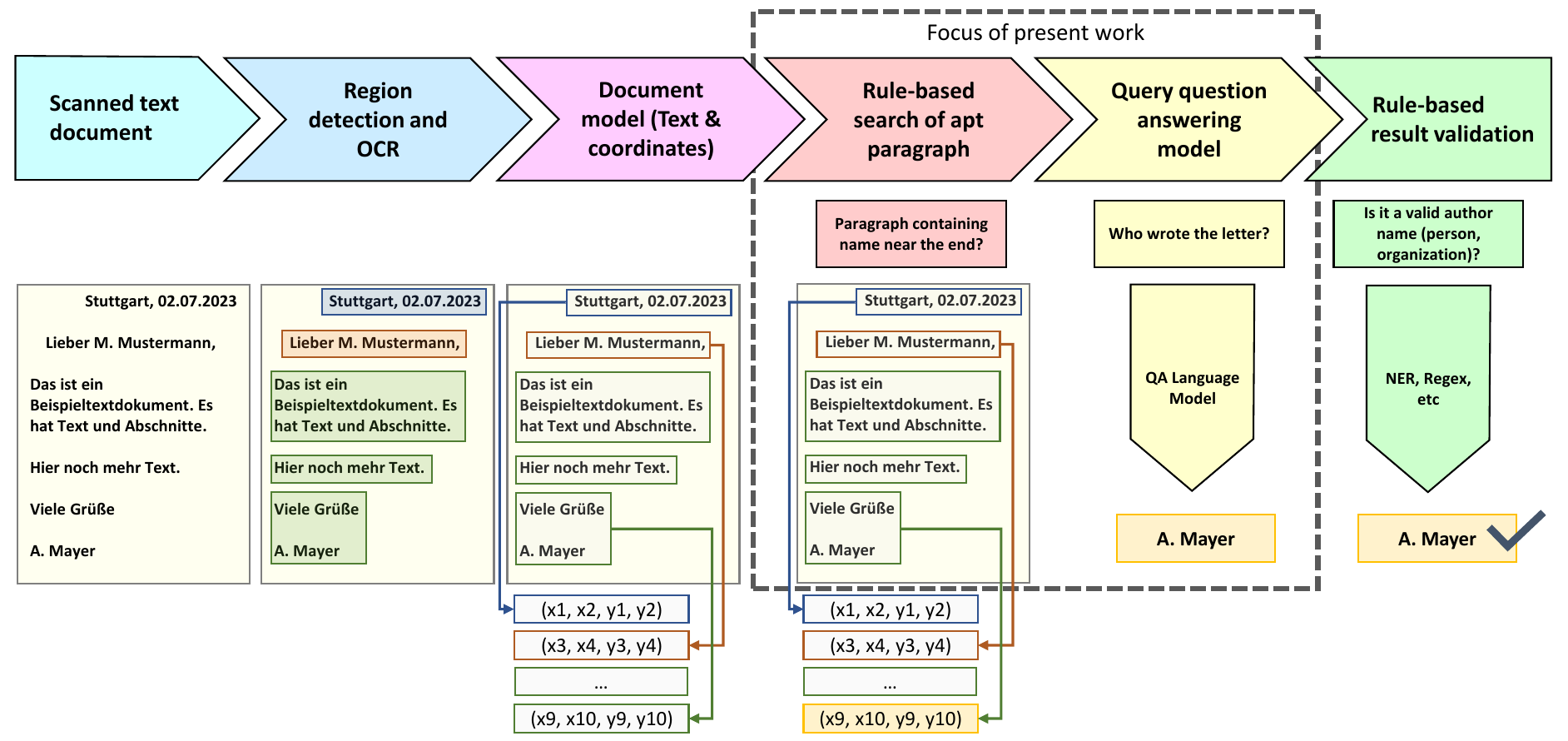, width=\textwidth}}
  \caption{Information Extraction Pipeline starting with visual image processing parts (page segmentation and OCR). Afterwards, the relevant features are extracted using extractive \acf{QA} models (focus of current work).}
  \label{fig:pipeline}
 \end{figure*}

The implemented pipeline includes multiple prepocessing and data reconstruction steps, the modules for the combined rule- and \ac{QA}-based information extraction and a final result evaluation. The architecture is shown in Figure~\ref{fig:pipeline}.

The \ac{IRS} combines the analysis of layout information (text position on the page, text size, column and tables, etc.) with textual analysis to narrow down and extract the relevant information for a given task.
The analysis pipeline works as follows:
\begin{enumerate}
    \item In a first step, a scanned text document (typically in PDF format) is provided as input to the framework and converted to raw image data.
    \item Using adapted region detection algorithms based on components of the German OCR-D project \cite{10.1145/3322905.3322917}, text blocks are detected and classified (categories can vary depending on the use case, but may include dates, sender or receiver address data, standard text paragraphs, table regions, image regions, etc.). \Ac{OCR} is performed to transform image to text data using optimized workflows based on Tesseract OCR \cite{tesseract}.
    \item The results are saved as an extended document model, containing the information to region, text content, and respective coordinates (on a region and character basis). 
    \item Depending on the use case and the usual length of input documents, the search scope may be restricted to the relevant text region or document page that is finally sent to the QA model. This is done using a rule-based approach, for example using keywords or headlines to help identify the proper candidate text regions.
    \item With the final search scope, the \ac{QA} model is then queried by providing the extracted text from the candidate regions and a suitable question targeting the information of interest.
    As output, the model returns the subsection of the candidate region with the highest probability for containing the answer.
    \item Finally, to avoid outputs that are clearly wrong (\eg numbers when asking for a name or vice-versa), a rule-based validation of the model answer is performed.
\end{enumerate}

The focus of this work lies on the 
the steps 3 and 4 and evaluating the trained \ac{QA} models for later integration into the larger \ac{IRS} framework,
as highlighted by the box in Figure~\ref{fig:pipeline}. 

\begin{figure*}[!t]
  \centering
   {\epsfig{file = 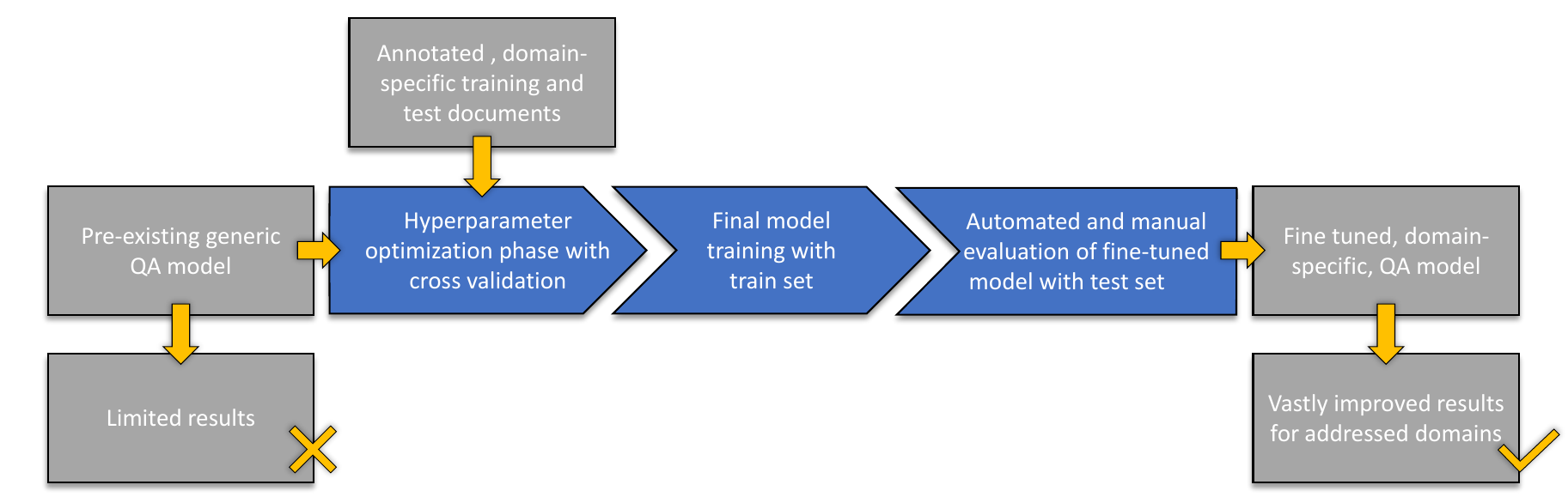,width=\textwidth}}
  \caption{Model Fine-tuning and Evaluation Process starting from a general German base QA model and using hyperparameter optimization with cross validation for before the final (best) model is trained and integrated for productive usage.}
  \label{fig:training}
\end{figure*}

\subsection{Fine-tuning and Evaluation Process }
\label{subsec:tuning}

For our task of domain specific QA fine-tuning we used the model \textit{gelectra-large-germanquad} that was pretrained on GermanQuAD~\cite{moller-etal-2021-germanquad}, a German variant of the popular SQuAD data set~\cite{SQuAD_Rajpurkar}. Both model and data set are provided by deepset and can be accessed via the common Huggingface API for inference and fine-tuning. 

To quantify the impact of domain-specific training, we constructed two distinct datasets, one targeting the medical domain and one for the insurance domain, comprising German language data. Each e dataset was enriched with pertinent information features relevant to the respective domain. In this context we chose the features to be different concerning properties like text length and complexity. In detail, we consider the two domains with following features:
\paragraph{Drug leaflet data set:}
    The \textit{leaflet} data set, which consists of 170 medication leaflet documents (which are freely available on many websites) with three QA pairs per document:
    \begin{itemize}
        \item \textbf{Ingredient}: the main active ingredient contained in the drug, e.g. \textit{Metoprololtartrat} 
        \item \textbf{Look}: The description of the drug appearance and optics, e.g. \textit{White, round pills}
        \item \textbf{Application}: The application scope of the drug, e.g. \textit{moderate pain and fever}
    \end{itemize}
\paragraph{Elemental damage report data set:}
    The \textit{report} data set, which consists of 47 elemental damage reports documents taken from one of our former projects in the insurance domain and coming with 2 QA pairs per document:
    \begin{itemize}
        \item \textbf{Damage Cause}: event description of what caused the damage, e.g. \textit{broken pipe due to rotted pipes in the floor} 
        \item \textbf{Assessor Name}: The name of the damage assessor who wrote the report, e.g. \textit{Manfred Bauer}
    \end{itemize}

\noindent We annotated all documents using the QA annotation tool Haystack\cite{haystack} with according questions asking for the specific entity of interest, e.g. \textit{What can the drug be applied for?} or \textit{What was the cause of the damage?}.

The further training process is described in Figure~\ref{fig:training}. In detail, since in this approach we only use data sets with limited amounts of samples, we used 5-fold cross validation splits of 80\% / 20\% for train and test to train several models in a grid search approach to find the optimal hyperparameter settings for both data sets, namely the values for epoch number, batch size, learning rate and doc stride. The latter one describes the amount of overlapping tokens when splitting up large documents into smaller chunks to comply with the maximum input size of the QA models (usually 512 tokens). Thus our script performs training and inference on smaller portions of the documents and collects merged predictions among the whole document to find the top confidence answer candidates during model queries.
 
We compare model performance before and after fine-tuning using automatically computable metrics described in Section~\ref{subsection:metrics} for settling the optimized hyperparameter configuration for training of the final model.
For training the final QA model with optimized hyperparameters we again used one single train / test split of 80\% train and 20\% test data, which was also evaluated with the additional manual expert metric presented in Section~\ref{subsection:metrics}. In the following we present our evaluation results and key findings for domain specific QA fine-tuning.

\begin{table*}[h]
\caption{Evaluation criteria for manual expert assessment for each data set and feature to extract. For many features, only a subset of the complete information is considered as sufficient regarding usefulness, e.g. if only last name of the assessor is detected}
\label{tab:evaluationcriteria}
\renewcommand{\arraystretch}{1.2}
\begin{tabular*}{\textwidth}{@{\extracolsep{\fill}}|l|l|m{10.0cm}|}
  \hline
  {\textbf{Data Set}} & {\textbf{Feature}} & {\textbf{Criteria for Answers to be rated as correct}} \\
  \hline
  Drug leaflets & Ingredient & All ingredients (there may be one or more) must be included in the answer, each with correct spelling \\
  \hline
  Drug leaflets & Look & Description of look (e.g. color and shape of pills) must be correct, details (e.g. notches) may be missing \\
  \hline
  Drug leaflets & Application & Description of application must be essentially correct, details may be missing \\
  \hline
  Damage reports & Damage cause & Description of cause must be essentially correct, different wording is acceptable; if there are several possible causes, one of these is sufficient \\
  \hline
  Damage reports & Assessor name & Last name must be exact, first name may be missing or abbreviated, title may be missing\\
  \hline
\end{tabular*}
\end{table*}

\section{\uppercase{Evaluation}}
\label{section:evaluation}

\subsection{Evaluation metrics}
\label{subsection:metrics}

Evaluating models in the context of \ac{NLP} is a complex task: statistically relevant results require a large number of labeled data points and thus an automated evaluation. At the same time, 
creating labeled data sets for specific \ac{NLP} tasks tend to be even more challenging than in other areas of supervised learning since natural language is fuzzy by nature and many different formulations can have the same meaning or represent a smooth transition between a correct and faulty statement.

Another difficulty is the fact that the correct answer to a question can appear multiple times in a document. For example, the name of the writer of an insurance claim assessment may appear on each page, and may be formatted in different ways ("John Doe" in the header, "J. Doe" in the text itself, "Mr. Doe" before the signature on the last page.) All these examples are correct answers to the question "Who wrote the document?" but report bad performance scores when evaluated with metrics that can not account for fuzziness in natural language, like most automatically evaluated scores.

To address these issues, we combined several evaluation metrics - a common practice when evaluating \ac{QA} models \cite{su-etal-2019-generalizing,drawehn2020}. For that, we formulate the objective in terms of a supervised learning problem. For a specific question $k$ from the set of all evaluated questions $q_k \in \mathcal{Q}$, we denote the labeled set of word vectors of correct answers from $N_k$ different annotators as $y^{(k)} = \bigl\{ y_1,\dots,y_{N_k} \bigr\}$ (without exact duplicates) and the response of a model to that question as \(\hat{y}^{(k)}\).
The metrics we used to evaluate our models are described below:

\begin{description}
\setlength{\itemsep}{5pt}
  \item[Manual Expert Assessment] As a ground-truth baseline, we evaluate a set of model answers manually, which although cumbersome and by definition not automated, is the only way to know if the answer provided by the \ac{QA} model indeed helps accomplish the task that the human end-user was interested in - this is the gold standard reference metric.
  
  To get evaluation values for our data sets, we put ourselves in the role of a user with common knowledge in the subject under consideration and rated all answers as correct that were at least partially correct and not misleading for the user. An overview of the rating criteria for the features in our data sets is given in Table~\ref{tab:evaluationcriteria}. Note that the thresholds for considering extracted information as correct resp. useful strongly depend on the type of feature: while short or even one-word entities like ingredient name leave little room for fuzzy extractions, for other features like look or assessor name it may be sufficient to only extract the most meaningful part of the feature (i.e. color and shape of the drug for look and the last name of the assessor. Other features like damage cause or drug application, which might be mentioned multiple times at different locations in the document context, are rated as correct if the found information is regarded as reasonable, complete and meaningful enough to answer to the question.
  Exact matches are always rated as correct.
  \item[Exact Match] $\mathcal{L}_{EM}$ measures the exact agreement of the model output with regard to the labeled answer(s) on a character basis after text normalization (\eg lower-case conversion, removal of control characters, etc.) for a question $k$ and is defined as
  \begin{equation*}
  \mathcal{L}_{EM}^{(k)} = \min\bigg\{1,\; \sum_{i=1}^{N_k} \mathbbold{1}\Big( \hat{y}^{(k)} = y_i \Big)\bigg\} ,
  \end{equation*}
  where $\mathbbold{1}$ is the indicator function. If the characters of the model's prediction exactly match the characters of (one of) the true answer(s), \ac{EM} returns $1$ and $0$ otherwise. This is a strict all-or-nothing metric, which means being off by a single character results in a score of $0$ for that question. The metric gives a good indication of the model performance when assessing against negative examples, where the answer to the provided question is not in the text. In this case, if the model predicts any text at all, it automatically receives a score of $0$ for that example.
  The metric is easy to check automatically, however, often insufficient for more complex answers.
  
  \item[Levenshtein] To measure the similarity between a true answer of a given question to its corresponding model output, and also account for the possibly very diverse responses to the same query in natural language, we use the \textit{Levenshtein} distance $\mathcal{L}_{Lev}$ \cite{Levenshtein66} as a character-based distance metric.
  Levenshtein measures the amount of operations (\ie insertion, deletion and substitution) that separate two strings of characters.
  
  \begin{table*}[t]
\vspace{-0.2cm}
\caption{Final hyperparameter configuration for fine-tuning experiments for each data set that have been determined during cross validation phase.}
\label{tab:parameters}
\renewcommand{\arraystretch}{1.1}
\begin{tabular*}{\textwidth}{@{\extracolsep{\fill}}|c|c|c|c|c|c|}
  \hline
  \textbf{Data set} & \textbf{Base Model} & \textbf{Epochs} &\textbf{Batch Size} & \textbf{Learning rate} & \textbf{Doc Stride} \\
  \hline
  Leaflets & deepset-gelectra-large-germanquad & 2 & 12 & 0.00001 & 128 \\
  \hline
  Reports & deepset-gelectra-large-germanquad & 5 & 12 & 0.00001 & 128 \\
  \hline
\end{tabular*}
\end{table*}

  \item[F1-score] Furthermore, we use the definition of \cite{SQuAD_Rajpurkar} to calculate the F1 score $\mathcal{L}_{F1}$ in the \ac{NLP} setting by computing the equal, \textit{word-wise} contribution between precision and recall, where precision is the ratio of the number of shared words to the total number of words in the prediction and recall is the ratio of the number of shared words to the total number of words in the ground truth~\cite{SQuAD_Rajpurkar,SQuAD2_Rajpurkar}:
  \begin{equation*}
  \mathcal{L}_{F1}^{(k)} = \frac{1}{N_k} \sum_{i=1}^{N_k} \frac{2}{\frac{|\mathcal{S}_{\hat{y}}^{(k)}|}{|\mathcal{S}_{y_i}^{(k)} \bigcap \mathcal{S}_{\hat{y}}^{(k)}|} + \frac{|\mathcal{S}_{y_i}^{(k)}|}{|\mathcal{S}_{y_i}^{(k)} \bigcap \mathcal{S}_{\hat{y}}^{(k)}|}}
  \end{equation*}
  Here \({\scriptstyle \mathcal{S}_{\hat{y}}^{(k)} }\) denotes the set of distinct words in the model prediction ${\scriptstyle \hat{y}^{(k)} }$ for a given question $k$, \({\scriptstyle \mathcal{S}_{y_i}^{(k)} }\) the word-set of one of the labeled answers $i$, and \(|\mathcal{S}_{\star}|\) the set size, i.e. number of unique elements (words) in the set.
  
  \item[ROUGE-L] We additionally calculate the \ac{ROUGE} metric $\mathcal{L}_{RGE}$ \cite{lin-2004-rouge}, a widely used scoring method to evaluate the summarization quality of a model for a given generated and one or more reference summaries. Specifically, we calculate the \ac{ROUGE}-L variant, denoted here as $\mathcal{L}_{RGE}$, which looks for the longest common subsequence (LCS) in the n-grams of two given sequences.
  In the extractive \ac{QA} setting, we can treat the model output and ground truth in the same way, since with the prior of a given question, the response is a de facto summary of the whole context.
  
  \item[Weighted average] Finally, we compute a weighted average $\mathcal{L}_{WA}$ of the above automated metrics $\mathcal{L}_{C}$ with $C = \{\text{EM} \mathrm{,\:} \text{Lev} \mathrm{,\:} \text{F1} \mathrm{,\:} \text{RGE}\}$ as a single score, that indicates the quality of the model response with regard to the different aspects of each individual metric:
  \begin{equation*}
  \mathcal{L}_{WA} = \dfrac{\sum_{l \in C} w_l \: \mathcal{L}_l}{\sum_{l \in C} w_l}
  \end{equation*}
  The weights $w_l$ are determined by a linear model trained on the $\mathcal{L}_{C}$'s and the expert assessment score as the label. To verify that the weights determined by this method are transferable to other \ac{QA} contexts, we train a regression model on the baseline and fine-tuned \ac{QA} models of each dataset and compare their deviation. 

  With this method we aim to create an automatically calculable metric that consists of a combination of individual scores and approximates the implicit criteria from human feedback to rate a model answer.
\end{description}

\noindent The calculation of the overall score $\mathcal{L}_{C}$ for each of the described metrics is done by averaging over all queries $\mathcal{Q}$ for one specific dataset, model and question type:
\begin{equation*}
  \mathcal{L}_{C} = \frac{1}{|\mathcal{Q}|}\sum_{k=1}^{|\mathcal{Q}|} \mathcal{L}_{C}^{(k)}
\end{equation*}

\subsection{Experimental Setup}

Following the fine-tuning approach described in Section \ref{subsec:tuning} we ended up with two final models trained with the configuration shown in Table \ref{tab:parameters}: one for the leaflet document use case and one for the damage report use case. We used 80\% of the data for training, the other 20\% were hold back as test sets for the final model evaluation, namely 35 leaflets and 10 report documents.

To measure the effect of our fine-tuning we compared model performances before and after the training process. Table \ref{tab:eval_results} lists the results for both data sets with according questions posed to the base and the fine tuned QA model using the metrics introduced in the previous Section \ref{subsection:metrics}. 

\begin{table*}[h]
\caption{Automated Evaluation of base and fine tuned models on medical leaflets and damage reports test sets.}
\label{tab:eval_results} 
\begin{tabular*}{\textwidth}{@{\extracolsep{\fill}}|l|l|l|P{1.8cm}|P{1.9cm}|P{0.8cm}|P{1.7cm}|P{1.2cm}|}
  \hline
  \textbf{Model} & \textbf{Dataset} & \textbf{Question} & \specialcell[t]{\textbf{Levenshtein}\\$\mathcal{L}_{Lev}$} & \specialcell[t]{\textbf{Exact Match}\\$\mathcal{L}_{EM}$} & \specialcell[t]{\textbf{F1}\\$\mathcal{L}_{F1}$} & \specialcell[t]{\textbf{ROUGE-L}\\$\mathcal{L}_{RGE}$} & \specialcell[t]{\textbf{Human}\\\textbf{Expert}} \\
  \specialrule{.2em}{.0em}{.0em}
  Base & Leaflets & Ingredient & 0.960 &	0.771 & 0.849 & 0.909 & 0.971 \\
  Fine-tuned & Leaflets & Ingredient & 0.985 &	0.914 &	0.941 & 0.959 &	1.000 \\
  \hline
  Base & Leaflets & Look & 0.611 &	0.147 &	0.452 & 0.468 &	0.529 \\
  Fine-tuned & Leaflets & Look & 0.710 &	0.206 &	0.657 & 0.678 &	0.824 \\
  \hline
  Base & Leaflets & Application & 0.563 &	0.030 &	0.434 & 0.436 &	0.758 \\
  Fine-tuned & Leaflets & Application & 0.761 & 0.212 & 0.694 & 0.713 & 0.909 \\
  \specialrule{.15em}{.0em}{.0em}
  Base & Reports & Damage Cause & 0.581 &	0.000 &	0.368 & 0.363 &	0.800  \\
  Fine-tuned & Reports & Damage Cause & 0.654 &	0.200	& 0.469 & 0.464  &	0.800 \\
  \hline
  Base & Reports & Assessor Name & 0.671 &	0.400 &	0.560 & 0.547 &	0.600  \\
  Fine-tuned & Reports & Assessor Name & 0.771 & 0.700 & 0.700 & 0.700 & 0.700 \\
  \hline
\end{tabular*}
\end{table*}

\subsection{Results and Discussion}

The outputs of the metric computations introduced in Section \ref{subsection:metrics} indicate a notable increase of model performance for the specific tasks, while the degree of improvements varies among the different data sets and questions with respect to the features of interest. 

For instance, with a score of 0.77 and an F1 score of about 0.85  the feature \textit{Ingredient} from the leaflet data set was already detected well by the base model (and got best overall scores). This might be due to the fact that the texts indicating this feature usually only consist of single and very specific words (like "Tamoxifencitrat") and additionally are announced by prominent keywords (like "The ingredient is..."), which makes it easy for a QA model to answer this question.

In contrast, the \textit{Damage Cause} feature, which is usually formed by one or several whole sentences with free formulation, seems to be the hardest to extract correctly due to its complexity. Nonetheless, also for this case we observe an increase of performance achieved by the QA fine-tuning process. Note that here also the base model already gave useful insights providing helpful information for finding the cause of the damage - even if this was not the originally labeled passage in the text (see human expert criteria in Table \ref{tab:evaluationcriteria}) -, which is also reflected in the comparatively higher Human Expert Assessment metric. 

The biggest improvement effect could be measured for the leaflet feature \textit{Look}: While the base model had difficulties to answer this question correctly (which might be also due to particularities in the way the question was formulated), the fine tuned model seems to have learned this feature very well, even with having little training samples available, which is indicated by a score increment of more than 0.25 for some of the metrics. 

\begin{figure*}[htb]
  \centering
  \begin{subfigure}[b]{0.49\textwidth}
    \includegraphics[width=\textwidth]{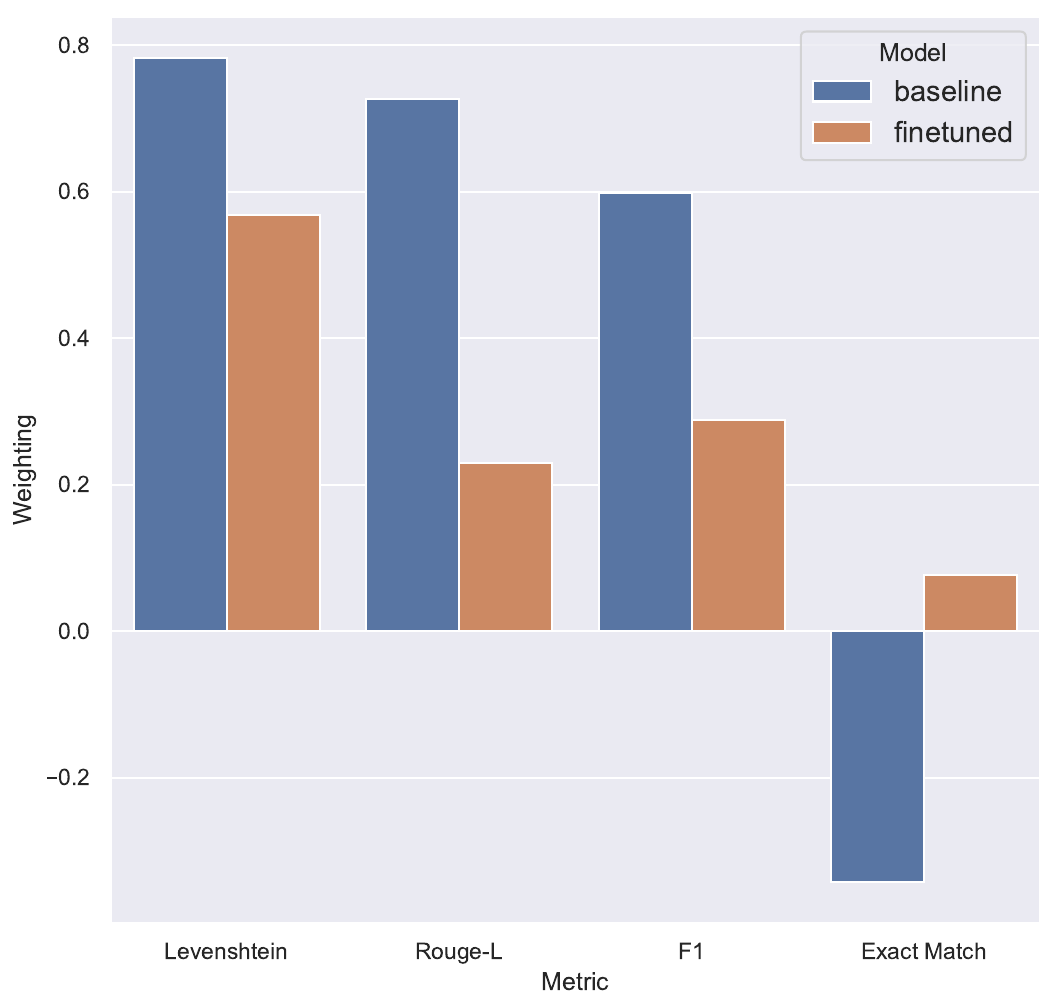}
    \caption{}
    \label{fig:coeffs_data}
  \end{subfigure}
  \hspace{4pt}
  \begin{subfigure}[b]{0.49\textwidth}
    \includegraphics[width=\textwidth]{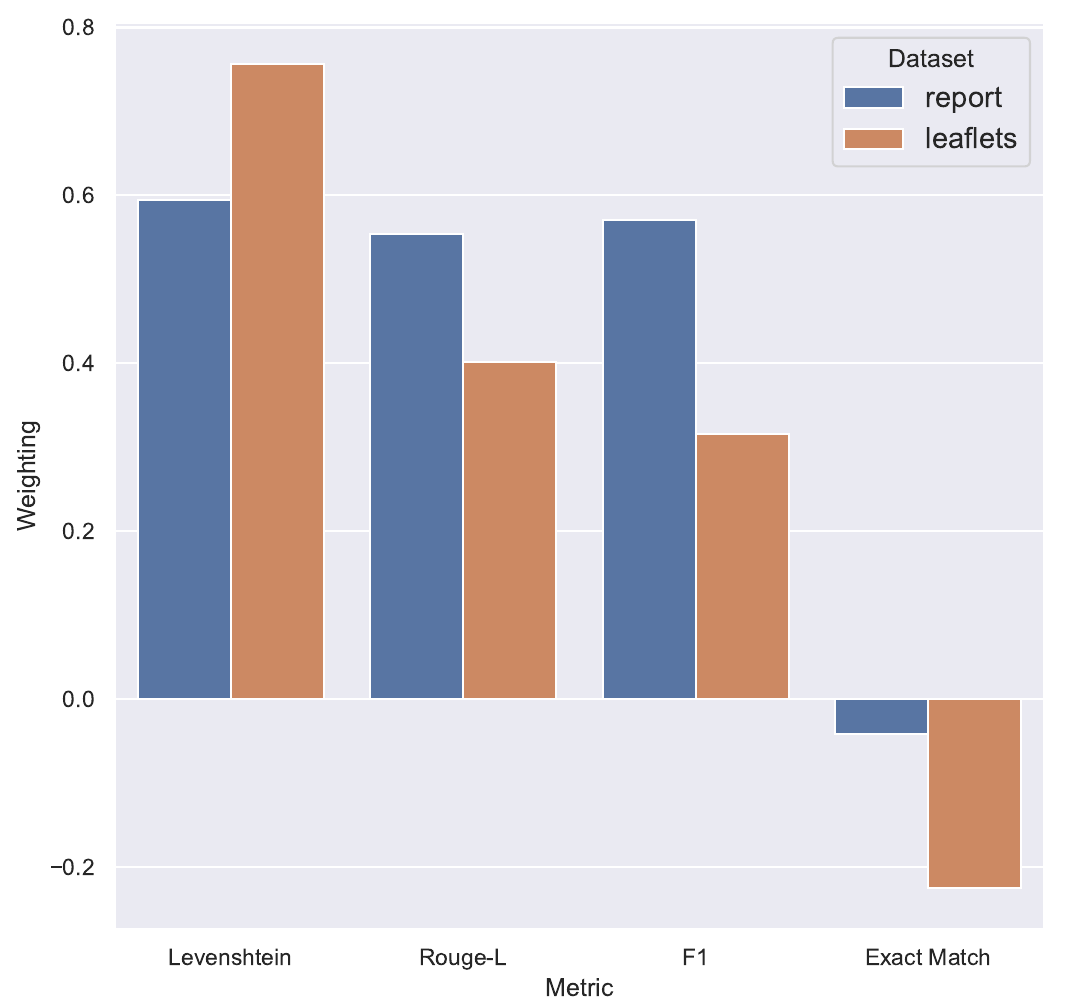}
    \caption{}
    \label{fig:coeffs_model}
  \end{subfigure}
  \caption{Weighting factors for the individual components $\mathcal{L}_{Lev}$, $\mathcal{L}_{RGE}$, $\mathcal{L}_{F1}$, $\mathcal{L}_{EM}$ of the weighted average metric. The behavior of the weights are compared between the applied models (a) and datasets (b) separately.}
  \label{fig:coeffs}
\end{figure*}

In general, the results of F1 and \ac{ROUGE}-L appear most similar to each other throughout all data sets and questions. Together with the Levenshtein, which plays to most important factor to approach the human metric through the weighted average score, they constitute results similar to those gained during the manual evaluation. In contrast, the \ac{EM} metric behaves totally different and does not seem to provide any clue about human result usefulness, a fact that is underlined by the outcomes of the weighted average score computation illustrated in Figures~\ref{fig:coeffs_data} and ~\ref{fig:coeffs_model}.

\subsection{Human Evaluation Score Approximation}
We train the linear model to predict the importance coefficients of the individual, automatically computable metrics from Section~\ref{subsection:metrics} to resemble the manual expert assessment score, which measures the helpfulness from a human perspective. 
The experiments show that the model is able to reconstruct the human scoring with a high accuracy of $93.87\%$ using $\mathcal{L}_{Lev}$, $\mathcal{L}_{RGE}, \mathcal{L}_{F1}$, and $\mathcal{L}_{EM}$ as features. The coefficient values of the linear model are used as the weights $w_l$ in the $\mathcal{L}_{WA}$ metric and shown in Figure~\ref{fig:coeffs}. A generalization of this approach over datasets and tasks from different domains could not be observed for our case. While the weighting factors for the \textit{reports} dataset are almost equally distributed between Levenshtein, \ac{ROUGE} and F1, with \ac{EM} basically being neglectable, for the \textit{leaflets} dataset a decaying importance of the individual metrics from Levenshtein to F1 can be observed, \ac{EM} even having a negative influence on the prediction. 
For both data sets the \ac{EM} metric is a poor factor in reconstructing the implicit aspects of what humans perceive as a useful answer, which is not very surprising, considering \ac{EM} as the hardest metric while humans still find an answer useful, even if some characters are missing or added to the model response.

In terms of derivation of the implicit rules for the human definition of helpfulness from a set of simple computable measures, we see metric behaviors that are in line with observations from \cite{schaeffer2023emergent}: for strict metrics like \ac{EM}, the baseline and (less severe) the fine-tuned model often produce much lower evaluation results than for the soft ones like F-measure or Levenshtein. 
This emphasizes that until a model becomes powerful enough to develop emergent abilities, strict metrics are normally less useful to catch the actual performance of the model for its use-case.

\section{\uppercase{Conclusion and future work}}
\label{section:conclusion}

In this paper, we showed that applying extractive \ac{QA} models for industrially relevant use cases of complex feature extraction leads to good performance for different kinds of domains, linguistic features and documents. Fine-tuning these \ac{QA} models makes significant improvement possible and helps to support or automate document analysis. Finally, we show that a weighted average over Levenshtein, ROUGE-L and F1 is a good approximation for manual human expert evaluation, whereas \ac{EM} is not. For future work, we want to further improve the results by tackling the following aspects:
\begin{itemize}
   \item Further fine-tuning of QA models, for example with larger data sets, to get even better accuracy for specific applications areas
   \item Prompt optimization and answer combination for queries with different wordings for the same question, for example "Who wrote the document?", "Who is the author of the document?" and "Which person wrote the document?" and use majority or another heuristic to decide on the final answer,
   \item Experimenting with multiple choice questions when applicable, for example "Was the author, John Doe or Max Mustermann?", as done in previous work \cite{jiang2021know}.
   \item Improvement of page segmentation and region detection to limit the query scope for the \ac{QA} model by feeding it only the most relevant parts of the text document for better response quality chances.
   \item Application of rule-based post-validation strategies for assuring quality and reliability of the feature predictions provided by the QA models.
   \item Investigation of multi-modal QA models that also take into account visual features like regions, boxes and page coordinates.
\end{itemize}

We further plan to include the best results and models as part of the document analysis pipeline of our industrial platform solution Aikido
\footnote{\url{https://www.digital.iao.fraunhofer.de/de/leistungen/KI/Aikido.html}}.

\bibliographystyle{apalike}
{\small
\bibliography{paper}}

\begin{thebibliography}{}

\bibitem[Bang et~al., 2023]{bang2023multitask}
Bang, Y., Cahyawijaya, S., Lee, N., Dai, W., Su, D., Wilie, B., Lovenia, H.,
  Ji, Z., Yu, T., Chung, W., Do, Q.~V., Xu, Y., and Fung, P. (2023).
\newblock A multitask, multilingual, multimodal evaluation of chatgpt on
  reasoning, hallucination, and interactivity.

\bibitem[{Cloudera Fast Forward Labs}, 2020]{SQuAD2}
{Cloudera Fast Forward Labs} (2020).
\newblock Squad2.0: The stanford question answering dataset.
\newblock \url{https://rajpurkar.github.io/SQuAD-explorer/}.
\newblock Accessed: 2022-12-12.

\bibitem[deepset, 2023]{haystack}
deepset (2023).

\bibitem[Drawehn et~al., 2020]{drawehn2020}
Drawehn, J., Blohm, M., Kintz, M., and Kochanowski, M. (2020).
\newblock Goal-based evaluation of text mining results in an industrial use
  case.
\newblock pages 183--191.

\bibitem[Engelbach et~al., 2022]{Engelbach_Klau_Drawehn_Kintz_2022}
Engelbach, M., Klau, D., Drawehn, J., and Kintz, M. (2022).
\newblock Combining deep learning and reasoning for address detection in
  unstructured text documents.

\bibitem[Glaese et~al., 2022]{Sparrow2022_Glaese}
Glaese, A., McAleese, N., Trebacz, M., Aslanides, J., Firoiu, V., Ewalds, T.,
  Rauh, M., Weidinger, L., Chadwick, M., Thacker, P., Campbell-Gillingham, L.,
  Uesato, J., Huang, P.-S., Comanescu, R., Yang, F., See, A., Dathathri, S.,
  Greig, R., Chen, C., Fritz, D., Elias, J.~S., Green, R., Mokrá, S.,
  Fernando, N., Wu, B., Foley, R., Young, S., Gabriel, I., Isaac, W., Mellor,
  J., Hassabis, D., Kavukcuoglu, K., Hendricks, L.~A., and Irving, G. (2022).
\newblock Improving alignment of dialogue agents via targeted human judgements.

\bibitem[Han et~al., 2021]{han2021cushlepor}
Han, L., Sorokina, I., Erofeev, G., and Gladkoff, S. (2021).
\newblock cushlepor: customising hlepor metric using optuna for higher
  agreement with human judgments or pre-trained language model labse.

\bibitem[Han et~al., 2013]{Han2013LanguageindependentMF}
Han, L., Wong, D.~F., Chao, L.~S., He, L., Lu, Y., Xing, J., and Zeng, X.
  (2013).
\newblock Language-independent model for machine translation evaluation with
  reinforced factors.
\newblock In {\em Machine Translation Summit}.

\bibitem[Jiang et~al., 2021]{jiang2021know}
Jiang, Z., Araki, J., Ding, H., and Neubig, G. (2021).
\newblock How can we know when language models know? on the calibration of
  language models for question answering.

\bibitem[Kahle et~al., 2017]{8270253}
Kahle, P., Colutto, S., Hackl, G., and Mühlberger, G. (2017).
\newblock Transkribus - a service platform for transcription, recognition and
  retrieval of historical documents.
\newblock In {\em 2017 14th IAPR International Conference on Document Analysis
  and Recognition (ICDAR)}, volume~04, pages 19--24.

\bibitem[Kintz et~al., 2020]{Kintz_Dukino_Blohm_Hanussek_2020}
Kintz, M., Dukino, C., Blohm, M., and Hanussek, M. (2020).
\newblock {M}ake your {C}ustomers {H}appy {A}gain. {AI} and {NLP} for a
  {C}ustomer {C}omplaint {M}anagement {P}latform.

\bibitem[Korbak et~al., 2023]{RLHP2023_Korbak}
Korbak, T., Shi, K., Chen, A., Bhalerao, R., Buckley, C.~L., Phang, J., Bowman,
  S.~R., and Perez, E. (2023).
\newblock Pretraining language models with human preferences.

\bibitem[Lambert et~al., 2022]{lambert2022rlhf}
Lambert, N., Castricato, L., von Werra, L., and Havrilla, A. (2022).
\newblock Illustrating reinforcement learning from human feedback (rlhf).
\newblock {\em Hugging Face Blog}.
\newblock https://huggingface.co/blog/rlhf.

\bibitem[Levenshtein, 1966]{Levenshtein66}
Levenshtein, V.~I. (1966).
\newblock Binary codes capable of correcting deletions, insertions, and
  reversals.
\newblock In {\em Soviet Physics Doklady, vol. 10, no. 8}, pages 707--710.

\bibitem[Lin, 2004]{lin-2004-rouge}
Lin, C.-Y. (2004).
\newblock {ROUGE}: A package for automatic evaluation of summaries.
\newblock In {\em Text Summarization Branches Out}, pages 74--81, Barcelona,
  Spain. Association for Computational Linguistics.

\bibitem[Microsoft,
  2022]{https://www.microsoft.com/en-us/research/project/document-ai/}
Microsoft (2022).
\newblock Document ai (intelligent document processing).
\newblock \url{https://www.microsoft.com/en-us/research/project/document-ai/}.
\newblock Accessed: 2022-12-20.

\bibitem[M{\"o}ller et~al., 2021]{moller-etal-2021-germanquad}
M{\"o}ller, T., Risch, J., and Pietsch, M. (2021).
\newblock {G}erman{Q}u{AD} and {G}erman{DPR}: Improving non-{E}nglish question
  answering and passage retrieval.
\newblock In {\em Proceedings of the 3rd Workshop on Machine Reading for
  Question Answering}, pages 42--50, Punta Cana, Dominican Republic.
  Association for Computational Linguistics.

\bibitem[Neudecker et~al., 2019]{10.1145/3322905.3322917}
Neudecker, C., Baierer, K., Federbusch, M., Boenig, M., W\"{u}rzner, K.-M.,
  Hartmann, V., and Herrmann, E. (2019).
\newblock Ocr-d: An end-to-end open source ocr framework for historical printed
  documents.
\newblock In {\em Proceedings of the 3rd International Conference on Digital
  Access to Textual Cultural Heritage}, DATeCH2019, page 53–58, New York, NY,
  USA. Association for Computing Machinery.

\bibitem[Ouyang et~al., 2022]{InstructGPT2022_Ouyang}
Ouyang, L., Wu, J., Jiang, X., Almeida, D., Wainwright, C., Mishkin, P., Zhang,
  C., Agarwal, S., Slama, K., Gray, A., Schulman, J., Hilton, J., Kelton, F.,
  Miller, L., Simens, M., Askell, A., Welinder, P., Christiano, P., Leike, J.,
  and Lowe, R. (2022).
\newblock Training language models to follow instructions with human feedback.
\newblock In Oh, A.~H., Agarwal, A., Belgrave, D., and Cho, K., editors, {\em
  Advances in Neural Information Processing Systems}.

\bibitem[Papineni et~al., 2002]{bleu2002Papineni}
Papineni, K., Roukos, S., Ward, T., and Zhu, W.-J. (2002).
\newblock Bleu: A method for automatic evaluation of machine translation.
\newblock In {\em Proceedings of the 40th Annual Meeting on Association for
  Computational Linguistics}, ACL '02, page 311–318, USA. Association for
  Computational Linguistics.

\bibitem[Pearce et~al., 2021]{DBLP:journals/corr/abs-2110-03142}
Pearce, K., Zhan, T., Komanduri, A., and Zhan, J. (2021).
\newblock A comparative study of transformer-based language models on
  extractive question answering.
\newblock {\em CoRR}, abs/2110.03142.

\bibitem[Rajpurkar et~al., 2018]{SQuAD2_Rajpurkar}
Rajpurkar, P., Jia, R., and Liang, P. (2018).
\newblock Know what you don't know: Unanswerable questions for squad.

\bibitem[Rajpurkar et~al., 2016]{SQuAD_Rajpurkar}
Rajpurkar, P., Zhang, J., Lopyrev, K., and Liang, P. (2016).
\newblock Squad: 100,000+ questions for machine comprehension of text.

\bibitem[Sanderson and Croft, 2012]{6182576}
Sanderson, M. and Croft, W.~B. (2012).
\newblock The history of information retrieval research.
\newblock {\em Proceedings of the IEEE}, 100(Special Centennial
  Issue):1444--1451.

\bibitem[Schaeffer et~al., 2023]{schaeffer2023emergent}
Schaeffer, R., Miranda, B., and Koyejo, S. (2023).
\newblock Are emergent abilities of large language models a mirage?

\bibitem[Smith, 2019]{tesseract}
Smith, R. (2019).
\newblock Tesseract ocr: an optical character recognition engine for various
  operating systems.
\newblock \url{https://github.com/tesseract-ocr/tesseract}.
\newblock Accessed: 2021-11-10.

\bibitem[Su et~al., 2019]{su-etal-2019-generalizing}
Su, D., Xu, Y., Winata, G.~I., Xu, P., Kim, H., Liu, Z., and Fung, P. (2019).
\newblock Generalizing question answering system with pre-trained language
  model fine-tuning.
\newblock In {\em Proceedings of the 2nd Workshop on Machine Reading for
  Question Answering}, pages 203--211, Hong Kong, China. Association for
  Computational Linguistics.

\bibitem[Tang et~al., 2022]{https://doi.org/10.48550/arxiv.2212.02623}
Tang, Z., Yang, Z., Wang, G., Fang, Y., Liu, Y., Zhu, C., Zeng, M., Zhang, C.,
  and Bansal, M. (2022).
\newblock Unifying vision, text, and layout for universal document processing.

\bibitem[Wolf et~al., 2020]{wolf-etal-2020-transformers}
Wolf, T., Debut, L., Sanh, V., Chaumond, J., Delangue, C., Moi, A., Cistac, P.,
  Rault, T., Louf, R., Funtowicz, M., Davison, J., Shleifer, S., von Platen,
  P., Ma, C., Jernite, Y., Plu, J., Xu, C., Scao, T.~L., Gugger, S., Drame, M.,
  Lhoest, Q., and Rush, A.~M. (2020).
\newblock Transformers: State-of-the-art natural language processing.
\newblock In {\em Proceedings of the 2020 Conference on Empirical Methods in
  Natural Language Processing: System Demonstrations}, pages 38--45, Online.
  Association for Computational Linguistics.

\bibitem[Xu et~al., 2021]{DBLP:journals/corr/abs-2110-06393}
Xu, P., Liang, D., Huang, Z., and Xiang, B. (2021).
\newblock Attention-guided generative models for extractive question answering.
\newblock {\em CoRR}, abs/2110.06393.

\bibitem[Zhang et~al., 2023]{zhang2023preliminary}
Zhang, J., Chen, Y., Niu, N., and Liu, C. (2023).
\newblock A preliminary evaluation of chatgpt in requirements information
  retrieval.

\bibitem[Ziegler et~al., 2019]{Ziegler2019FineTuningLM}
Ziegler, D.~M., Stiennon, N., Wu, J., Brown, T.~B., Radford, A., Amodei, D.,
  Christiano, P., and Irving, G. (2019).
\newblock Fine-tuning language models from human preferences.
\newblock {\em ArXiv}, abs/1909.08593.

\end{thebibliography}



\end{document}